\pdfoutput=1
\documentclass[runningheads]{llncs}
\usepackage{graphicx}
\usepackage{booktabs}
\usepackage{array}
\newcolumntype{L}[1]{>{\raggedright\let\newline\\\arraybackslash\hspace{0pt}}m{#1}}
\newcolumntype{C}[1]{>{\centering\let\newline\\\arraybackslash\hspace{0pt}}m{#1}}
\newcolumntype{R}[1]{>{\raggedleft\let\newline\\\arraybackslash\hspace{0pt}}m{#1}}
\begin{document}
\title{Learning Deformable Point Set Registration with Regularized Dynamic Graph CNNs for Large Lung Motion in COPD Patients}
\titlerunning{Deformable Point Set Registration with Regularized DGCNNs}
%
\author{Lasse Hansen \and
Doris Dittmer \and
Mattias P. Heinrich}

\authorrunning{L. Hansen et al.}
%
\institute{Institute of Medical Informatics, University of L\"ubeck, Germany\\
\email{\{hansen, dittmer, heinrich\}@imi.uni-luebeck.de}}
\maketitle              
\begin{abstract}
Deformable registration continues to be one of the key challenges in medical image analysis. While iconic registration methods have started to benefit from the recent advances in medical deep learning, the same does not yet apply for the registration of point sets, e.g. registration based on surfaces, keypoints or landmarks. This is mainly due to the restriction of the convolution operator in modern CNNs to densely gridded input. However, with the newly developed methods from the field of geometric deep learning suitable tools are now emerging, which enable powerful analysis of medical data on irregular domains. In this work, we present a new method that enables the learning of regularized feature descriptors with dynamic graph CNNs. By incorporating the learned geometric features as prior probabilities into the well-established coherent point drift (CPD) algorithm, formulated as differentiable network layer, we establish an end-to-end framework for robust registration of two point sets. Our approach is evaluated on the challenging task of aligning keypoints extracted from lung CT scans in inhale and exhale states with large deformations and without any additional intensity information. Our results indicate that the inherent geometric structure of the extracted keypoints is sufficient to establish descriptive point features, which yield a significantly improved performance and robustness of our registration framework.

\end{abstract}
\section{Introduction and Related Work}
Registration, i.e. determining a spatial transformation that aligns two images or point sets, is a fundamental task in medical image and shape analysis and a prerequisite for numerous clinical applications. It is widely used for image-guided intervention, motion compensation in radiation therapy, atlas-based segmentation or monitoring of disease progression. Non-rigid registration is ill-posed and thus a non-convex optimization problem with a very high number of degrees of freedom. In addition, the medical domain poses particular challenges on the registration task, e.g. non-linear intensity differences in multi-modal images or high inter-patient variations in anatomical shape and appearance.

\textbf{Iconic registration: }Voxel-based intensity-driven medical image registration has been an active area of research, which can e.g. be solved using discrete \cite{glocker2008dense} optimization of a similarity metric and a regularization constraint on the smoothness of the deformation field. Data driven deep learning methods based on convolutional neural networks (CNNs),  have only recently been used in the field of medical image registration. In \cite{de2019deep} an iconic and unsupervised learning approach is introduced that learns features to drive a registration and replaces the iterative optimization with a feed-forward CNN. While achieving impressive runtimes of under a second on a GPU the accuracy for CT lung motion estimation is inferior to conventional methods. Weak supervision in the form of landmarks or multi-label segmentations was used in the CNN framework of \cite{hu2018weakly}, where the similarity measure is based on the alignment of the registered labels.

\textbf{Geometric registration:} To capture large deformations, e.g. present in intra-patient inhale-exhale examinations of COPD patients \cite{castillo2013reference} or vessel-guided brain shift compensation \cite{bayer2018intraoperative}, geometric registration models - based on keypoints or surfaces - offer a promising solution. Point-based registration has not yet profited from the advantages of deep feature learning due to the restriction of conventional CNNs to densely gridded input. Many current geometric methods (e.g. \cite{bayer2018intraoperative} and \cite{ravikumar2019generalised}) are based on the well-established coherent point drift (CPD) algorithm~\cite{myronenko2010point}. In addition to 3D coordinates, they incorporate further image or segmentation-derived features, such as point orientations or scalar fractional anisotropy (FA) values \cite{ravikumar2019generalised}.

\textbf{Deep geometric learning: }While these hand-crafted features clearly improved on the results of the CPD, recent methods from the field of geometric deep learning~\cite{bronstein2017geometric} would enable a data-driven feature extraction directly from point sets. The PointNet framework \cite{qi2017pointnet} was one of the first approaches to apply deep learning methods to unordered point sets. A limitation of the approach is that is does not consider local neighborhood information, which was adressed in \cite{wang2018dgcnn} by dynamically building a k-nearest-neighbour graph on the point set and thus also enabling feature propagation along edges in that graph. Combining convolutional feature learning with a differentiable and robustly regularized fitting process has first been proposed for multi-camera scene reconstruction in \cite{brachmann2017dsac} (DSAC), but has so far been limited to rigid alignment.

\textbf{Large deformation lung registration: }Both iconic and geometric approaches have often been found to yield relative large residual errors for large motion lung registration (forced inhale-to-exhale): e.g. 4.68 mm for the discrete optimization algorithm in~\cite{glocker2008dense} applied to the DIR-lab COPD data \cite{castillo2013reference} and 3.61 mm (on the inhale-exhale pairs of the EMPIRE10 challenge) for \cite{ehrhardt2010automatic}, which used both keypoint- and intensity-based information. Learning the alignment of such difficult data appears to be so far impossible with intensity-driven CNN approaches that already struggle with more shallow breathing in 4D-CT \cite{de2019deep}. Thus being able to directly match vessel- and airway trees based on geometric features alone can provide a valuable pre-alignment for further intensity-based registration (cf.~\cite{heinrich2015estimating}) or be directly used in clinical applications to perform atlas-based labelling of anatomical segments and branchpoints for physiological studies \cite{tschirren2005matching}.

\subsection{Contributions}
Our work contributes two important steps towards data-driven point set registration that enables the incorporation of deep feature learning into a regularized CPD fitting algorithm. First, we utilize dynamic graph CNNs \cite{wang2018dgcnn} in an auxiliary metric learning task to establish robust correspondences between a moving and a fixed point set. These learned features are shown to yield an improved modeling of prior probabilities in the CPD algorithm. Since all operations of the CPD algorithm are differentiable, we secondly show that it is possible to further optimize the parameters of the feature extraction network directly on the registration task. To evaluate our method we register keypoints extracted from inhale and exhale states in lung CT-scans from the challenging DIR-Lab COPD dataset \cite{castillo2013reference} showing the general feasibility of a deep learning point set registration framework in an end-to-end manner and with only geometric information.

\section{Methods}
\label{sec:methods}

\begin{figure}[t]
\includegraphics[width=\textwidth]{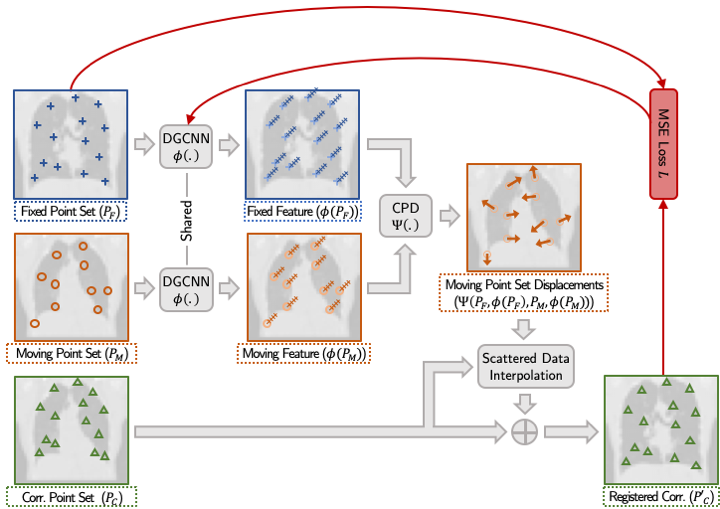}
\caption{Illustration of our proposed method for supervised non-rigid point set registration. While we investigate the problem of 3D registration, here, point sets are depicted in two dimensions for simplicity. Also, point sets are underlaid with coronal lung CT slices as visualization aides. No image information is used in our registration pipeline.}
\label{fig:idea}
\end{figure}

In this section, we introduce our proposed method for deformable point set registration with deeply learned features. Figure \ref{fig:idea} summarizes the methods general idea. Input to our method are the fixed point set $P_F$ and the moving point set $P_M$. While we make no assumptions on the number of points or correspondences in the input point sets, we assume a further set of keypoint correspondences with $P_F$ for the supervised learning task, which is denoted as $P_{C}$. We compute geometric features from $P_F$ and $P_M$ with a shared dynamic graph CNN (DGCNN \cite{wang2018dgcnn})~$\phi$. The spatial positions together with the extracted descriptors are input to the feature based CPD algorithm that produces displacement vectors for all points in $P_M$. We then employ thin-plate splines (TPS) \cite{bookstein1989principal} as a scattered data interpolation method to compute the displacements for $P_{C}$, which yields the transformed point set $P^{'}_{C}$. Finally, we can compute the mean squared error (MSE) of the Euclidean distance between correspondences in $P_F$ and $P^{'}_{C}$ as a loss $L$ for the optimization of the feature extraction network $\phi$. In the following, we describe the descriptor learning with the DGCNN as well as the extensions to the CPD algorithm to exploit point features as prior probabilities.

\subsection{Descriptor Learning on Point Sets with Dynamic Graph CNNs}

\begin{figure}[t]
\centering
\includegraphics[width=0.78\textwidth]{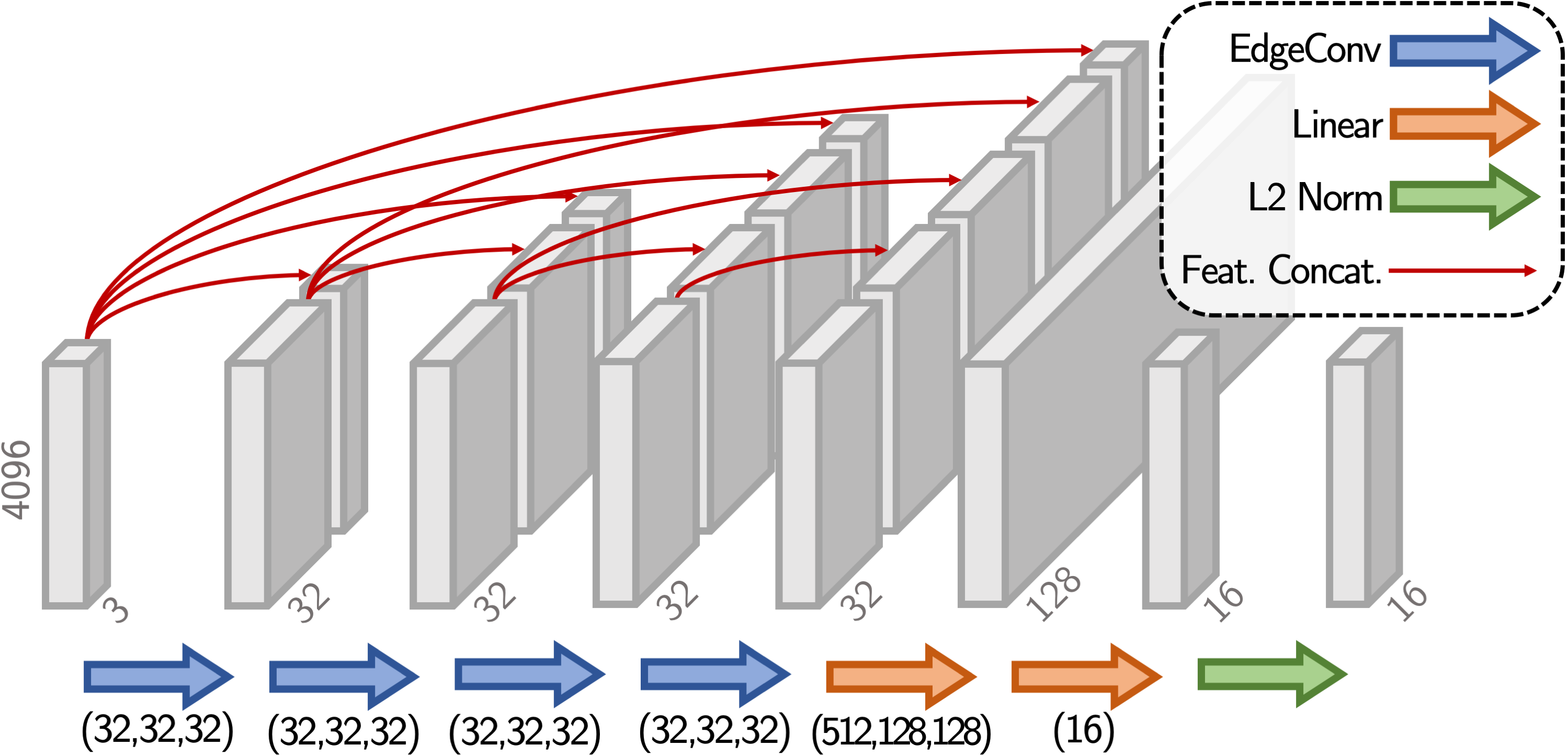}
\caption{Proposed network architecture for geometric feature extraction from the fixed and moving point set. Input is a three-dimensional point set and the network computes a 16-dimensional geometric descriptor for each of the 4096 points. The number of layer neurons for each operation is specified in the corresponding brackets.}
\label{fig:architecture}
\end{figure}

Our proposed network architecture for geometric feature extraction is illustrated in Figure \ref{fig:architecture}. A key component is the edge convolution introduced in \cite{wang2018dgcnn}, that dynamically builds a k-Nearest-Neighbor (kNN) graph from the points in the input feature space and then aggregates information from neighbouring points to output a final feature map. We employ several edge convolutions with DenseNet style feature concatenation to efficiently capture both local and global geometry. The final feature descriptor is obtained by fully connected layers that reduce the point information to a given dimensionality. We restrict the output descriptor space by $L_2$ normalization to enable constant parametrization of subsequent operations in the registration pipeline which stabilizes network training. To establish robust initial correspondences between the moving and fixed point set the model is pretrained in an auxiliary metric learning task using a triplet loss.

\subsection{Feature-based Coherent Point Drift}

The CPD algorithm formulates the alignment of two point sets as a probability density estimation problem. The points in the moving point set $P_M$ are described as centroids of gaussian mixture models (GMMs) and are fitted to the points in the fixed point set $P_F$ by maximizing the likelihood. To find the displacements for $P_M$ the Expectation Maximization (EM) algorithm is used, where in the E-step point correspondence probabilities $\mathbf{C}$ are computed and in the M-step the displacement vectors are updated. We incorporate the learned geometric feature descriptors $\phi(P_F)$ and $\phi(P_M)$ as additional prior probabilites with
\begin{equation}
\mathbf{C}(P_F, \phi(P_F), P_M, \phi(P_M)) = \mathbf{C_{pos}}(P_F, P_M) + \alpha \cdot \mathbf{C_{feat}}(\phi(P_F), \phi(P_M)),
\end{equation}
where $\mathbf{C_{pos}}$ denotes the spatial point correspondence described in \cite{myronenko2010point}, $\alpha$ is a trade-off and scaling parameter and
\begin{equation}
\label{eq:c_feat}
\mathbf{C_{feat_{mn}}}(\phi(P_F)_{n}, \phi(P_M)_{m}) = \exp(-\frac{1}{2\cdot\rho^2} \left\| \phi(P_F)_{n}-\phi(P_M)_{m}\right\|^2)
\end{equation}
with $n=1\ldots N$ and $m=1\ldots M$. $N$ and $M$ denote the number of points in $P_F$ and $P_M$, respectively. In addition to the parameter $\rho$ in (\ref{eq:c_feat}), that controls the width of the Gaussian, the CPD algorithm includes three more free parameters: $w$, $\lambda$ and $\beta$. Parameter $w$ models the amount of noise and outliers in the point sets, while parameters $\lambda$ and $\beta$ control the smoothness of the deformation field.

\section{Experiments}
Registering the fully inflated to exhaled lungs is considered one of the most demanding tasks in medical image registration, which is important for analyzing e.g. local ventilation defects in COPD patients. We use the DIR-Lab COPD data set~\cite{castillo2013reference} with 10 inhale-exhale pairs of 3D CT scans for all our experiments. The thorax volumes are resampled to isotropic voxel-sizes of $1$~mm and a few thousands keypoints are extracted from inner lung structures with the Foerstner operator. Automatic correspondences to supervise the learning of our DGCNN are established using the discrete and intensity-based registration algorithm of \cite{heinrich2015estimating}, which has an accuracy of $\approx$1 mm. In all experiments, no CT-based intensity information is used and all processing relies entirely on the geometric keypoint locations.

In our first experiment, we learn point descriptors directly in a supervised metric learning task. Therefore, a triplet loss is employed forcing feature similarity between corresponding keypoint regions in point set pairs. The inhale and exhale point set form the positive pair, while points from the permuted exhale point set yield as negative examples. These learned features can be directly used in a kNN registration. We then investigate the combination of spatial positions and learned descriptors in the feature-based CPD algorithm. Finally, in our concluding experiment, the feature network is trained in an end-to-end manner as described in Section \ref{sec:methods} to further optimize the pretrained geometric features.

\textbf{Implementation details:}
Due to the limited number of instances in the used dataset we perform a leave-one-out validation, where we evaluate on one inhale and exhale point set and train our network with the remaining nine pairs. During training we use farthest point sampling to obtain $4096$ points from the inhale and exhale point set, respectively. Each evaluation is run ten times and results are averaged to account for the effect of the sampling step. The employed network parameters are specified in Figure \ref{fig:architecture}. For the CPD algorithm ($250$ iterations) we use following parameters: $\alpha=0.05$, $\rho=0.5$ $w=0.1$, $\lambda=5$ and $\beta=1$. For the end-to-end training we relax parameters $\rho$ and $\beta$ to $0.25$ and $0.5$, respectively, to allow for further optimization of input features.

\section{Results and Discussion}
\begin{figure}[t]
\centering
\includegraphics[width=\textwidth]{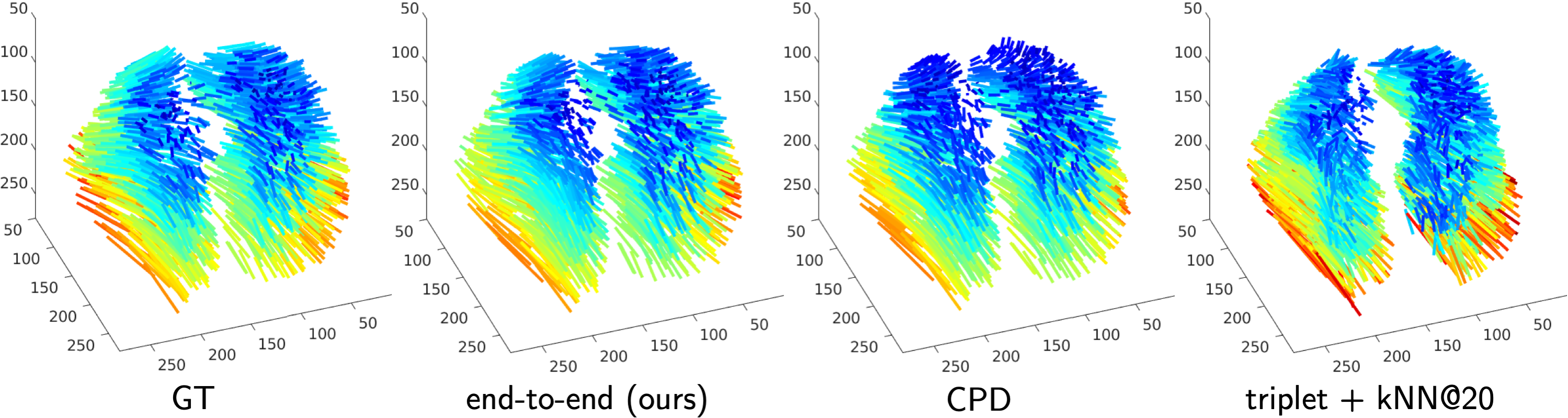}
\caption{Qualitative results in terms of 3D motion vectors on test case \#5. The magnitude is color coded from blue (small motion) to red (large motion).}
\label{fig:results}
\end{figure}

\begin{table}[t]
\centering
\caption{Results for the 10 inhale and exhale CT scan pairs of the DIR-Lab COPD data set~\cite{castillo2013reference}. The mean target registration error (TRE) in mm is computed on the 300 expert annotated landmark pairs per case. The $p$-values are obtained by a rank-sum test over all 3000 landmark errors with respect to our best performing approach.}
\label{tab:results}
\renewcommand{\arraystretch}{0.7}
\begin{tabular}{C{1.15cm}|C{1.5cm}|c|C{1.7cm}|C{1.5cm}|C{1.7cm}|C{1.7cm}}
  \toprule
  \bfseries Case \# & initial & center-aligned & triplet + kNN@20 & CPD \cite{myronenko2010point} & triplet + CPD (ours) & end-to-end (ours)\\
  \midrule
  \bfseries 1 & 26.3 & 17.8 & 8.1 & 5.5 & 4.2 & \bfseries 3.4 \\
  \bfseries 2 & 21.8 & 14.7 & 15.6 & \bfseries 8.4 & 9.3 & 8.9 \\
  \bfseries 3 & 12.6 & 10.6 & 6.4 & 2.7 & 2.5 & \bfseries 2.4 \\
  \bfseries 4 & 29.6 & 19.0 & 8.3 & 4.8 & 3.4 & \bfseries 3.2 \\
  \bfseries 5 & 30.1 & 18.4 & 7.8 & 8.4 & 5.2 &  \bfseries4.6 \\
  \bfseries 6 & 28.5 & 16.2 & 7.5 & 14.0 & 5.1 & \bfseries 4.3 \\
  \bfseries 7 & 21.6 & 10.2 & 6.3 & 3.0 & 2.6 & \bfseries 2.5 \\
  \bfseries 8 & 26.5 & 17.4 & 6.3 & 6.8 & 4.3 & \bfseries 3.9 \\
  \bfseries 9 & 14.9 & 14.1 & 9.0 & 3.5 & \bfseries 3.1 & 3.6 \\
  \bfseries 10 & 21.8 & 19.6 & 14.9 & \bfseries 7.4 & 7.5 & \bfseries 7.4 \\
  \midrule
  \bfseries mean & 23.4 & 15.7 & 9.0 & 6.4 & 4.7 & \bfseries 4.3 \\
  \bfseries std & 11.9 & 7.0 & 5.5 & 5.2 & 4.1 & \bfseries 3.6 \\
  \midrule
  \bfseries $p$-val & $<10^{-4}$ & $<10^{-4}$ &  $<10^{-4}$ & $<10^{-4}$ & $1.4\cdot10^{-2}$ & -\\
  \bottomrule
\end{tabular}
\end{table}

Qualitative results are shown in Figure \ref{fig:results} where our approach demonstrates a good trade-off between the very smooth motion of the CPD and the potential for large correspondences of the features from triplet-learning. Our quantitative results that are evaluated on 300 independent expert landmark pairs for each patient demonstrate that registering the point clouds directly with CPD (3D coordinates as input) yield a relatively large target registration error (TRE) of 6.4$\pm$5.2 mm (see Table~\ref{tab:results}). Employing kNN registration based on a DGCNN trained with keypoint correspondences to extract geometric features without regularization is still inferior with a TRE of 9.0$\pm$5.5 mm highlighting the challenges of this point-based registration task and the difficulties of addressing the deformable alignment with one-to-one correspondence search. Combining the geometric features of a pre-trained DGCNN with the regularizing CPD that is extended to use 19-dimensional inputs (16 features + 3 coordinates) yields a substantial improvement over each individual method with a TRE of 4.7$\pm$4.1 mm. Finally, using end-to-end learning to back-propagate the regularized alignment errors through the iterative point drift layers to further improve the feature learning shows another small but significant improvement to 4.3$\pm$3.6 mm. These alignment errors cannot be directly compared to the large variety of image- and feature-based registration algorithms that reached 3.6 mm \cite{ehrhardt2010automatic}, 4.7 mm \cite{glocker2008dense} or 1.1 mm \cite{heinrich2015estimating} for similar datasets, but were based on intensity information, while our comparison is restricted to purely geometric approaches without intensity. In addition, a better outcome would be expected by extending the keypoint extraction to focus on vessel- or airway-based nodes and to include anatomical tree-based edges in the graph model. Nevertheless, the results clearly showed that our models are already able to directly learn semantic geometric features in a data-driven manner based on the inherent correspondence information.

\section{Conclusion}
We have presented a new method for deformable point set registration that learns geometric features from irregular point sets using a dynamic graph CNN (DGCNN) together with a regularizing and fully differentiable high-dimensional coherent point drift (CPD) model. Our results clearly indicate that geometric feature learning, even from relatively uninformative point clouds, is possible with DGCNNs and can be further enhanced when incorporating the CPD model into the optimization. Evaluated on challenging inhale-exhale lung registration of COPD patients we achieve an improvement of 2.1 mm over the classical CPD method and are competitive with many classical image-based registration algorithms despite the fact that no intensity information is used. In addition to these encouraging findings, we believe that alternative regularization models to the CPD, that require fewer iteration steps could have potential to further improve this approach. In future works, many more applications, e.g. surface point shape alignment and analysis, could benefit from deep point registration.

\bibliographystyle{splncs04}
\bibliography{glmi19hansen-bibliography}

\begin{thebibliography}{10}
\providecommand{\url}[1]{\texttt{#1}}
\providecommand{\urlprefix}{URL }
\providecommand{\doi}[1]{https://doi.org/#1}

\bibitem{bayer2018intraoperative}
Bayer, S., Ravikumar, N., Strumia, M., Tong, X., Gao, Y., Ostermeier, M.,
  Fahrig, R., Maier, A.: Intraoperative brain shift compensation using a hybrid
  mixture model. In: MICCAI. pp. 116--124 (2018)

\bibitem{bookstein1989principal}
Bookstein, F.L.: Principal warps: Thin-plate splines and the decomposition of
  deformations. TPAMI  \textbf{11}(6),  567--585 (1989)

\bibitem{brachmann2017dsac}
Brachmann, E., Krull, A., Nowozin, S., Shotton, J., Michel, F., Gumhold, S.,
  Rother, C.: Dsac-differentiable ransac for camera localization. In: CVPR. pp.
  6684--6692 (2017)

\bibitem{bronstein2017geometric}
Bronstein, M.M., Bruna, J., LeCun, Y., Szlam, A., Vandergheynst, P.: Geometric
  deep learning: going beyond euclidean data. IEEE Signal Processing Magazine
  \textbf{34}(4),  18--42 (2017)

\bibitem{castillo2013reference}
Castillo, R., Castillo, E., Fuentes, D., Ahmad, M., Wood, A.M., Ludwig, M.S.,
  Guerrero, T.: A reference dataset for deformable image registration spatial
  accuracy evaluation using the copdgene study archive. Physics in Medicine \&
  Biology  \textbf{58}(9), ~2861 (2013)

\bibitem{ehrhardt2010automatic}
Ehrhardt, J., Werner, R., Schmidt-Richberg, A., Handels, H.: Automatic landmark
  detection and non-linear landmark-and surface-based registration of lung ct
  images. Medical Image Analysis for the Clinic-A Grand Challenge, MICCAI
  \textbf{2010},  165--174 (2010)

\bibitem{glocker2008dense}
Glocker, B., Komodakis, N., Tziritas, G., Navab, N., Paragios, N.: Dense image
  registration through mrfs and efficient linear programming. Medical image
  analysis  \textbf{12}(6),  731--741 (2008)

\bibitem{heinrich2015estimating}
Heinrich, M.P., Handels, H., Simpson, I.J.: Estimating large lung motion in
  copd patients by symmetric regularised correspondence fields. In: MICCAI. pp.
  338--345 (2015)

\bibitem{hu2018weakly}
Hu, Y., Modat, M., Gibson, E., Li, W., Ghavami, N., Bonmati, E., Wang, G.,
  Bandula, S., Moore, C.M., Emberton, M., et~al.: Weakly-supervised
  convolutional neural networks for multimodal image registration. Medical
  Image analysis  \textbf{49},  1--13 (2018)

\bibitem{myronenko2010point}
Myronenko, A., Song, X.: Point set registration: Coherent point drift. TPAMI
  \textbf{32}(12),  2262--2275 (2010)

\bibitem{qi2017pointnet}
Qi, C.R., Su, H., Mo, K., Guibas, L.J.: Pointnet: Deep learning on point sets
  for 3d classification and segmentation. In: Proceedings of the Conference on
  Computer Vision and Pattern Recognition. pp. 652--660 (2017)

\bibitem{ravikumar2019generalised}
Ravikumar, N., Gooya, A., Beltrachini, L., Frangi, A.F., Taylor, Z.A.:
  Generalised coherent point drift for group-wise multi-dimensional analysis of
  diffusion brain mri data. Medical image analysis  \textbf{53},  47 -- 63
  (2019)

\bibitem{tschirren2005matching}
Tschirren, J., McLennan, G., Pal{\'a}gyi, K., Hoffman, E.A., Sonka, M.:
  Matching and anatomical labeling of human airway tree. TMI  \textbf{24}(12),
  1540--1547 (2005)

\bibitem{de2019deep}
de~Vos, B.D., Berendsen, F.F., Viergever, M.A., Sokooti, H., Staring, M.,
  I{\v{s}}gum, I.: A deep learning framework for unsupervised affine and
  deformable image registration. Medical image analysis  \textbf{52},  128--143
  (2019)

\bibitem{wang2018dgcnn}
Wang, Y., Sun, Y., Liu, Z., Sarma, S.E., Bronstein, M.M., Solomon, J.M.:
  Dynamic graph cnn for learning on point clouds. arXiv preprint
  arXiv:1801.07829  (2018)

\end{thebibliography}

\end{document}